# Detection of Bleeding in Wireless Capsule Endoscopy Images Using Range Ratio Color

Amer A. Al-Rahayfeh and Abdelshakour A. Abuzneid

Computer Science and Engineering Department
University Of Bridgeport, Bridgeport, CT 06601, USA
{ aalrahay, abuzneid}@bridgeport.edu

**ABSTRACT**

*Wireless Capsule Endoscopy (WCE) is device to detect abnormalities in colon,esophagus,small intestinal and stomach, to distinguish bleeding in WCE images from non bleeding is a hard job by human reviewing and very time consuming. Consequently, automation for classifying bleeding frames not only will expedite the process but will reduce the burden on the doctors. Using the purity of the red color we can detect the Bleeding areas in WCE images. But, we could find various intensity of red color values in different parts of the small intestinal ,so it is not enough to depend on the red color feature alone. We select RGB(Red,Green,Blue) because it takes raw level values and it is easy to use. In this paper we will put range ratio color for each of R,G,and B. Therefore, we divide each image into multiple pixels and apply the range ratio color condition for each pixel. Then we count the number of the pixels that achieved our condition. If the number of pixels grater than zero, then the frame is classified as a bleeding type. Otherwise, it is a non-bleeding. Our experimental results show that this method could achieve a very high accuracy in detecting bleeding images for the different parts of the small intestinal*

## 1. INTRODUCTION

A human digestive system consists of a series of several different organs including the esophagus, stomach, small in- testinal (i.e., duodenum, jejunum, and ileum) and colon. Standard endoscopy has been playing a very important role as a diagnostic tool for the digestive tract. For example, various endoscopies such as gastroscopy, push enteroscopy colonoscopy have been used for the visualization of digestive system. However, all methods mentioned above are limited in viewing small intestine. To address the problem, Wire- less Capsule Endoscopy (WCE) was ¯first proposed in 2000, which integrates wireless transmission with image and video technology [6, 9, 14, 15, 16, 17,18].

The Wireless Capsule Endoscopy (WCE) is a non-invasive technique that enables the visualization of the small bowel mucosa for diagnosing purposes. The WCE is swallowed by the patient and it is passively propelled by peristalsis. The most common indication for capsule endoscopy is for evaluation of obscure gastrointestinal bleeding. Early systematic studies suggest that it is a very effective diagnostic tool in patients with Crohn's disease; moreover, it is expected that WCE would offer a clinical benefit in terms of small bowel cancer survival [1,18,19].

As shown in Figure 1 and figure 2, wireless capsule endoscopy, measuring 26mm × 11mm, is a pill-shaped device which consists of a short-focal-length CMOS camera, light source, battery and radio transmitter. We first introduce how it works briefly. After a WCE is swallowed by a patient who has a diet for about 12 hours, this little device propelled by peristalsis starts to work and record the images while moving forward along the digestive tract. Meanwhile, the images recorded by the camera are sent out wirelessly to a special recorder attached to the waist. This process continues for about eight hours until the WCE battery ends. Finally, all the image data in the special recorder are downloaded into a personal computer or a computer workstation, and physicians can view the images and analyze potential sources of different diseases in the gastrointestina  (GI) tract. It should be noted that the diagnosis process exerted by physicians is very time-consuming due to the large amount of video, so the diagnosis is not a real-time process. This situation paves a potential way for off-line post processing and computer aided diagnosis. The WCE was approved by U.S. Food and Drug Administration (FDA) in 2001, and it has been reported that this new technology shows great value in evaluating gastrointestinal bleeding, Crohn's disease, ulcer and other diseases existed in the digestive tract [2,18].

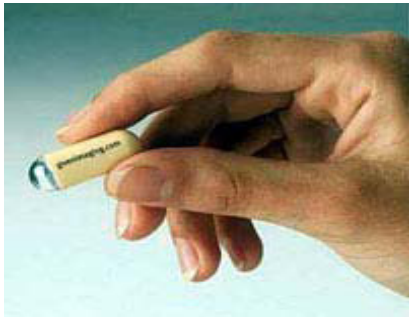

Figure 1 . Wireless capsule endoscopy[2]

There still remains some improvement for the WCE, although this new technology shows great advantages over traditional examination techniques. One problem associated with this technology is that it would take a long period of time for physicians to inspect the large number of images it produced. There are about 50,000 images in total per examination for one patient, and it costs an experienced clinician about two hours on average to review and analyze all the video data [2]. Figure 2 illustrate eight sample images of healthy regions and bleeding lesions of the Gastrointestinal (GI) tract.

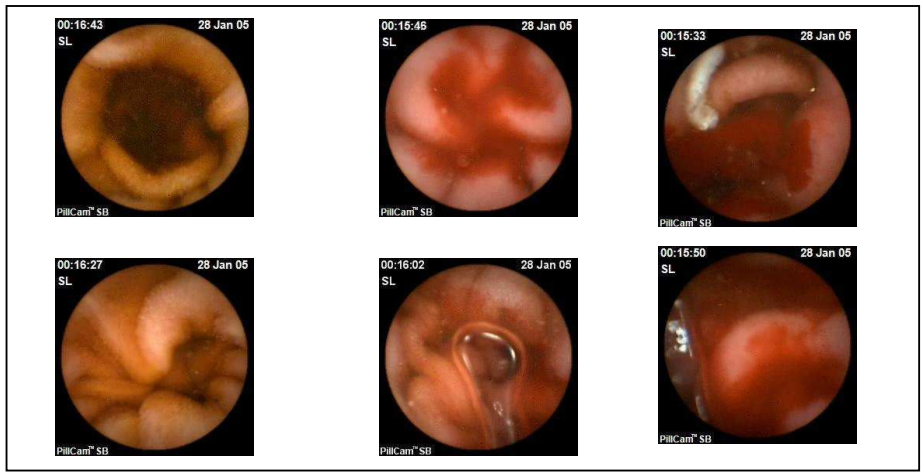

Figure 2. Video frames from CE videos that show normal (top row) and bleeding lesions (bottom row) [3].

Besides, abnormalities in the GI tract may be present in only one or two frames of the video, so they might be missed by physicians due to oversight sometimes. Moreover, there may be some abnormalities that cannot be detected by the naked eyes because of their size, color, texture and distribution. Furthermore, different clinicians may have different findings when come to the same image data. All these problems motivate the researchers to develop reliable and uniform assisting approaches to reduce the great burden of the physicians. However, it should be admitted that this goal is very challenging because the true features associated with the diseases are not exactly known. Moreover, different diseases have totally different symptoms in the digestive tract. Even the same disease shows great variations in color and shape [2].

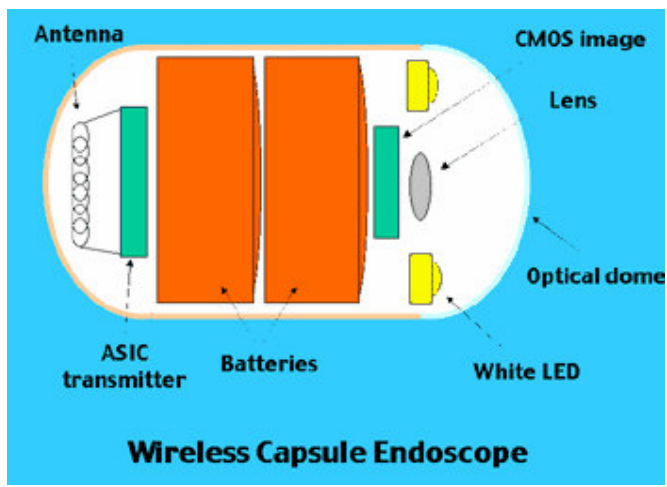

Figure 3. Components diagram of a WCE [2]

In this paper we focus on classifying the WCE video frames as either bleeding or none bleeding. The main feature used for characterizing bleeding areas is the color feature, since the intensity of the red color component differs for different parts of the GI tract .Using the Red color feature alone doesn't produce accurate results and fails in identifying bleeding images for different parts of the GI tract. Therefore, setting a range value for the shall produce better results.

This paper is organized as follows. In part 2 we give a brief description of some related work for detection algorithms. In part 3 we will show how our method works. Results and discussion will be discussed In part 4 . Finally part 5 concludes our work.

## 2. RELATED WORK

### 2.1 DEFINITIONS

Blood-based abnormalities in the small bowel are characterized in three categories: a) Bleeding, b) Angioectasia and c) Erythema. Such a categorization is important in order to comprehend the problem efficiently.

**A. Bleeding**

Bleeding is defined as the flow of blood from a ruptured blood vessel into the digest tract. Among malignant tumors, leiomyosarcoma is most commonly associated with bleeding. Bleeding with adenocarcinoma is less frequent, and rarely occurs with carcinoid tumors [11]. In a multicenter research by Pennazio et aI, [12], 29 of the 60 patients underwent push enteroscopy in addition to capsule endoscopy. Among those 29 patients, push enteroscopy detected a source of bleeding in 8 patients (280/0), whereas capsule endoscopy detected a source of bleeding in 17 patients (590/0) [10].

**B. Angioectasia**

In medical bibliography angioectasia is also referred to as arteriovenous malformations. It is the most common abnormality accounting for obscure gastrointestinal bleeding, seen in 210/0-53% of patients who undergo capsule endoscopy [13]. This makes their detection an important task. They occur more frequently with increasing age and can be identified at endoscopy as spiderlike lesions [10].

**C. Erythema**

In general, erythema is defined as skin redness caused by capillary congestion. In the small intestine erythema multiform is strongly related with various abnormalities (i.e. Croehn's disease). It is an acute, self-limiting, inflammatory skin eruption. Its redness is lesser than that of angioectasia [10].

Many papers in WCE image processing have recently appeared. Therefore, there is an increasing research interest in automatically discriminate abnormal tissues. Color feature is the most popular bleeding detection method. In [3], a two-steps process is proposed; the first step discriminates images with bleeding, using a block-based color saturation method; the second one refines the initial classification and increases its reliability using a pixel-based saturation-luminance analysis.

In [4], the authors proposed a two-stage approach to detect bleeding using  based on the support vector machine and  adaptive color histogram. In [5] the authors measure the usefulness of MPEG-7 for detection of a variety of events, such as bleeding, ulcers and polyps.

In [6] The authors proposed a method using color distribution to discriminate stomach, intestine and colon tissue. An interesting way of selecting the MPEG-7 visual descriptors as the feature extractor to do detection for several diseases such as ulcers and bleeding in the gastrointestinal tract was advanced in [5].

Bashar et al. [7] proposed an automatic method to differentiate informative from non-informative frames by constructing a color feature space for classification. Daniel Barbosa et al. [8] used texture feature which are extracted through wavelet transform and statistical estimation to detect small bowel tumors in WCE images.

In [9], the authors used clotted blood for bleeding. Although their technique aims to detect bleeding regions, it eventually detects blood regions. In their work the 2D color distributions show a significant overlapping of non-blood and blood pixels, though they claim the opposite. In addition, as stated above, a simple examination of the bleeding pixels and non-bleeding pixels can prove that their values are identical in many cases. Dismissing dark pixels from the image doesn't necessary imply removal of non-blood pixels. There are cases when the illumination of the WCE is low making bleeding regions darker. Their methodology dismisses small detected blood regions. It is needless to mention of the existence of small-sized bleeding regions. However, the use of Expectation Maximization clustering and Bayesian Information Criterion are very promising tools, offering an automatic clustering and a statistical approach to this difficult problem.

In [10] the authors proposed applying first color adaptation to the RGB images~ which offers image enhancement but the lack of sufficient results and statistics do not justify the efficiency of their methodology. In addition as mentioned before, the R component in WCE images is spread almost at all pixels at high values.

## 3. SIMULATION ENVIRONMENT

In this section we present the simulation results of  WCE images to be categorized as either bleeding or non-bleeding .

### 3.1. Simulation Platform

We will use C- sharp ( C# ) which  is a programming language which most directly reflects on the underlying Common Language Infrastructure (CLI).

### 3.2. Experiment Environment

Medical image analysis often requires the application of several image analysis techniques, exploiting different image features, to be used in conjunction, in order to address the various relevant aspects of the anatomically and pathologically meaningful regions. The proposed method for detecting gastrointestinal bleeding regions is divided into two main steps. The first step provides an efficient discrimination of the input videos that contain bleeding characteristics (bleeding) from those that do not correspond to bleeding (non-bleeding). The second step proceeds with a further evaluation of the bleeding images and verifying if the initial classification really corresponds to active bleeding patterns [3] .

To characterize the contractions in WCE videos, we tested several different features. The Color is  the main feature used to distinguish up- normal discoloration of the GI tissues. But, using the purity of the red color feature alone didn't produce accurate results and failed in identifying bleeding images.

In this paper we use range- ratio color  for bleeding. We first start by using the purity of the bleeding color then we use a range- ratio for each of R, G, and B.  In our analysis  we will divide the image into pixels like Figure 2.

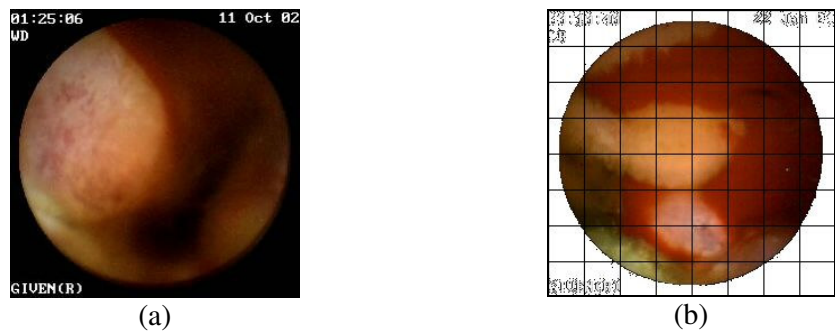

(a)                                    (b)

Figure 4. WCE image before and after divide into pixels[2]

Second, we start investigating pixel- by- pixel through applying by apply the purity of the red color (R=255,G=0,B=0).We use the following algorithm :

```
for (int i = 0; i < height; i++)
for (int j = 0; j < width; j++)
              {
         R = Pixels[i, j, 0];
         G = Pixels[i, j, 1];
         B = Pixels[i, j, 2];

if (R == 255 && G == 0 && B == 0)
              {
            ++c;
              }
            else
              {
            c1++;
              }
              }
          if (c > 0)
System.Console.WriteLine( "  Yes bleeding");
            else
System.Console.WriteLine( "  No bleeding");
```

Figure 5. Using the purity of the bleeding color

```
for (int i = 0; i < height; i++)
for (int j = 0; j < width; j++)
              {
         R = Pixels[i, j, 0];
         G = Pixels[i, j, 1];
         B = Pixels[i, j, 2];

If(R >= 75 && R < 128 && G <= 25 && G >= 14 && B <= 15 && B >=0)              {
            ++c;
              }
            else
              {
            c1++;
              }
              }
          if (c > 0)
System.Console.WriteLine( "  Yes bleeding");

            else
System.Console.WriteLine( "  No bleeding");
```

Figure 6. using a range- ratio-color

## 4. RESULTS AND DISCUSSION

To test our algorithm we used 100 different WCE images taken for different parts of the GI tract. Half of those images are classified visually by a real specialist as bleeding and the other half are non- bleeding. We used these images for checking the bleeding classification method which we proposed. We applied the first algorithm in images while the condition is (R=255,G=0,B=0)(i.e. Purity of RED value ), then we count the number of pixels that met our condition. If the number of pixels grater than zero, then the frame is classified as a bleeding type. Otherwise, it is a non-bleeding. We repeated the previous scenario by using a different condition which is (R >= 75 and R < 128, G <= 25 and G >= 14, B <= 15 and B >=0) .

The results are shown in table 1, Figure 7 and Figure 8. Negative number corresponds to prediction of non-bleeding image and positive number corresponds bleeding. It can be clearly seen that when we based our classification on Purity of RED value alone the total accuracy was only 48%. But when we used range- ratio - color feature we obtained an overall accuracy of 98% .This method the problem of first-false-positive prediction and hence gives a better accuracy.

Table 1: Results of bleeding classification method

| Classification | Number of bleeding Prediction | Number of none bleeding Prediction | Total Correct Prediction | Over all Accuracy |
|---|---|---|---|---|
| Purity of RED | 2 | 98 | 48 | 48% |
| Range Ratios Color | 52 | 48 | 98 | 98% |

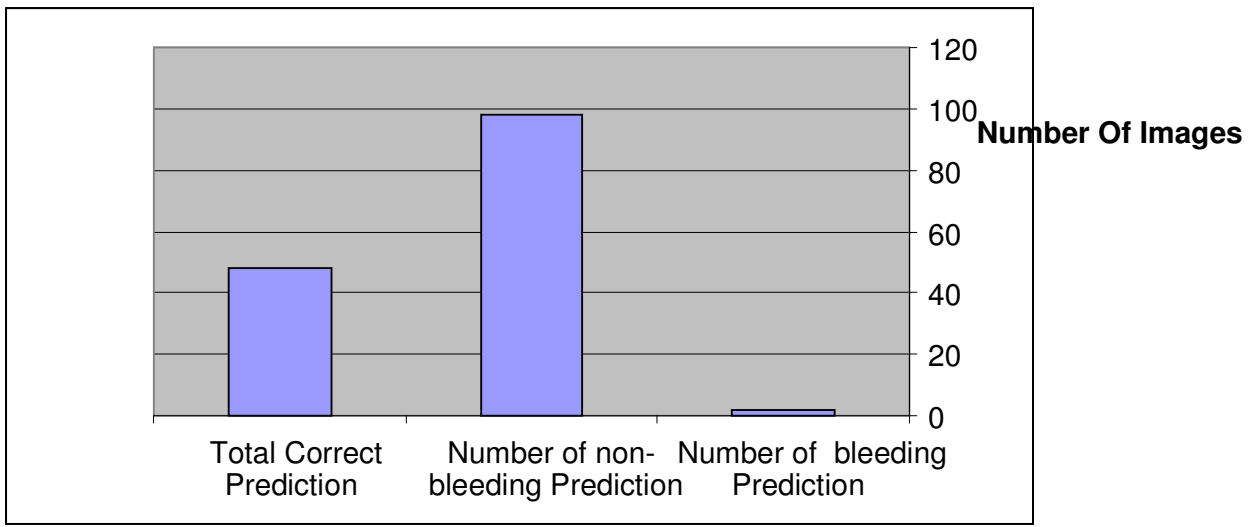

Figure 7. Using Purity of RED Value

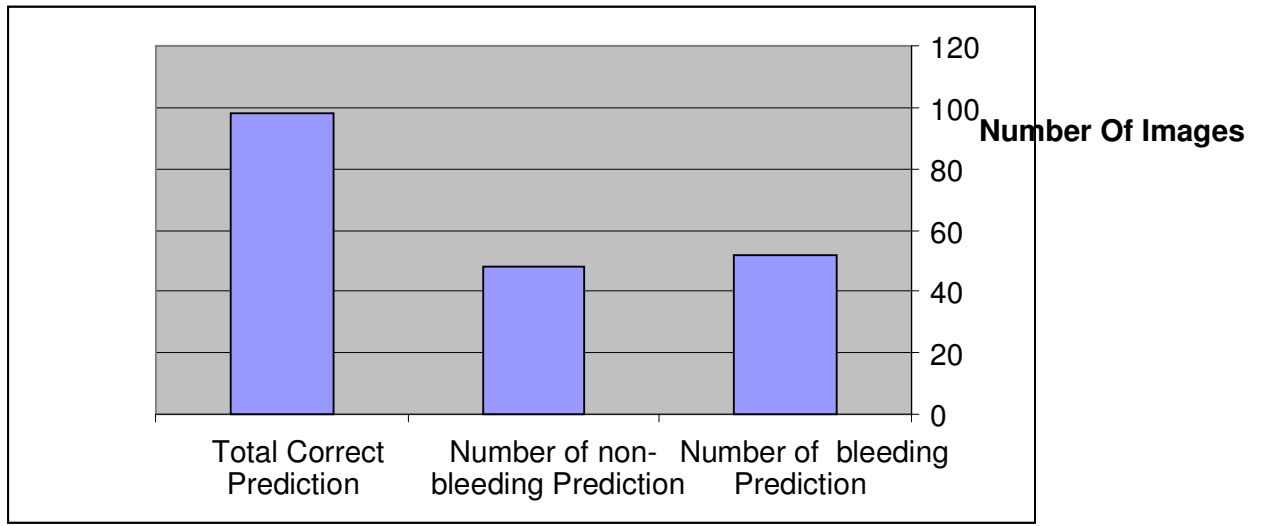

Figure 8. Using Range –Ratios-Color

## 5. CONCLUSIONS

In most images Bleeding is characterized by the red color value. Human eyes are mostly sensitive to brightness and lightness of the color not the color itself so the degree of brightness or illumination incurred by the light source associated with capsule affects the degree of how the red color is perceived and analyzed.

An novel method to differentiate between bleeding and non-bleeding images was proposed in this work. Our method is mainly based on dividing the WCE images into pixels then applying measures based on the purity of the red color and the range ratio-color-components. . It can be clearly seen that when we based our classification on Purity of RED value alone the total accuracy was only 48%. But when we used range-ratio-color feature we obtained an overall accuracy of 98% which corrects the problem of first false positive prediction and hence gives a better accuracy If we compare our method with some other complex algorithms we will find it is simple but gave a high accuracy in classifying as bleeding or non-bleeding.

**Authors**

Amer Al-Rahayfeh is a Ph.D. student of Computer Science and Engineering at the University of Bridgeport. He received a B.S. in Computer Science from Mutah University and an M.S.in Computer Information Systems from The Arab Academy for Banking and Financial in 2002 and 2004. He worked as an Instructor at The Arab Academy for Banking and Financial (2006-2008), Amer current research interests are in multimedia database systems.

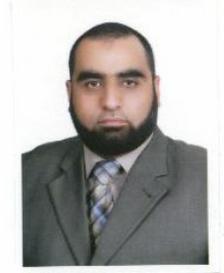

Abdelshakour Abuzneid has received his BS degree in Computer Engineering and Control from Yarmouk University and MS degree in Computer Engineering from the University of Bridgeport in May 2007.
Currently he is pursuing his PhD in Computer Science & Engineering from the University of Bridgeport. His research interest is in Data / computer / wireless / mobile communications. He has published few journal and conference papers.

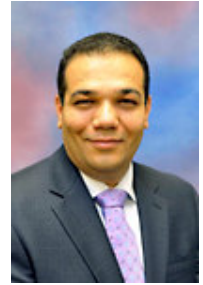